\documentclass[
]{ceurart}

\usepackage{listings}
\usepackage{xcolor}
\newcommand{\deepseek}{\textsc{DeepSeek-V3.1}\xspace}
\newcommand{\mistralsmall}{%
  \textsc{Mistral-Small-3.2-24B}\xspace}
\newcommand{\gemma}{\textsc{Gemma-3-27B-it}\xspace}
\newcommand{\qwen}[1]{\textsc{Qwen3-VL-#1-Instruct}\xspace}
\newcommand{\qwenfam}{\textsc{Qwen3-VL}\xspace}
\newcommand{\grobid}{\textsc{GROBID}\xspace}

\newcommand{\cex}{\textsc{CEX}\xspace}
\newcommand{\excite}{\textsc{EXCITE}\xspace}
\newcommand{\linkbook}{\textsc{LinkedBooks}\xspace}

\begin{document}

%%
%% Rights management information.
%% CC-BY is default license.
\copyrightyear{2026}
\copyrightclause{Copyright for this paper by its authors.
  Use permitted under Creative Commons License Attribution 4.0
  International (CC BY 4.0).}

%%
%% This command is for the conference information
\conference{SCOLIA 2026 — The Second International Workshop on Scholarly Information Access}

%%
%% The "title" command
\title{Benchmarking Large Language Models on Reference Extraction and Parsing in the Social Sciences and Humanities}

% \tnotemark[1]
% \tnotetext[1]{You can use this document as the template for preparing your
%   publication. We recommend using the latest version of the ceurart style.}

%%
%% The "author" command and its associated commands are used to define
%% the authors and their affiliations.
\author[1]{Yurui Zhu}[%
orcid=0009-0002-1429-1139, %TODO
email=yurui.zhu@odoma.ch,
]
\cormark[1]
\address[1]{\href{www.odoma.ch}{Odoma LLC}}

\author[1]{Giovanni Colavizza}[
orcid=0000-0002-9806-084X,
email=giovanni.colavizza@odoma.ch,
url=, %TODO
]
  
\author[1]{Matteo Romanello}[
orcid=0000-0002-7406-6286, %TODO
email=matteo.romanello@odoma.ch,
url=, %TODO
]
% \fnmark[1]
% \address[1]

% \author[4]{Manfred Jeusfeld}[%
% orcid=0000-0002-9421-8566,
% email=Manfred.Jeusfeld@acm.org,
% url=http://conceptbase.sourceforge.net/mjf/,
% ]
% \fnmark[1]
% \address[4]{University of Skövde, Högskolevägen 1, 541 28 Skövde, Sweden}

% %% Footnotes
\cortext[1]{Corresponding author.}
% \fntext[1]{These authors contributed equally.}

%%
%% The abstract is a short summary of the work to be presented in the
%% article.
\begin{abstract}
Bibliographic reference extraction and parsing are foundational for citation indexing, linking, and downstream scholarly knowledge-graph construction. However, most established evaluations focus on clean, English, end-of-document bibliographies, and therefore underrepresent the Social Sciences and Humanities (SSH), where citations are frequently multilingual, embedded in footnotes, abbreviated, and shaped by heterogeneous historical conventions. We present a unified benchmark that targets these SSH-realistic conditions across three complementary datasets: \cex (English journal articles spanning multiple disciplines), \excite (German/English documents with end-section, footnote-only, and mixed regimes), and \linkbook (humanities references with strong stylistic variation and multilinguality). We evaluate three tasks of increasing difficulty—reference extraction, reference parsing, and end-to-end document parsing—under a schema-constrained setup that enables direct comparison between a strong supervised pipeline baseline (\grobid) and contemporary LLMs (\deepseek, \mistralsmall, \gemma, and \qwenfam (4B–32B variants)). Across datasets, extraction largely saturates beyond a moderate capability threshold, while parsing and end-to-end parsing remain the primary bottlenecks due to structured-output brittleness under noisy layouts. We further show that lightweight LoRA adaptation yields consistent gains—especially on SSH-heavy benchmarks—and that segmentation/pipelining can substantially improve robustness. Finally, we argue for hybrid deployment via routing: leveraging \grobid for well-structured, in-distribution PDFs while escalating multilingual and footnote-heavy documents to task-adapted LLMs.
\end{abstract}

%%
%% Keywords. The author(s) should pick words that accurately describe
%% the work being presented. Separate the keywords with commas.

%%
%% This command processes the author and affiliation and title
%% information and builds the first part of the formatted document.
\maketitle

\section{Introduction}
Reliable citation processing is a prerequisite for large-scale scholarly infrastructure, as it facilitates the creation of curated bibliographies, citation graphs and citation indexes. 
% In practice, reference extraction and parsing remain fragile once we move beyond ``laboratory'' conditions. 
In practice, reference extraction and parsing remain fragile and challenging in Social Sciences and Humanities (SSH) literature, where citations often appear in footnotes, rely on abbreviations and ellipsis, mix languages within a document, and follow style conventions that vary across periods, venues, and national traditions. These characteristics hinder both (i) \emph{reference extraction} (detecting and delimiting reference spans within full text) and (ii) \emph{reference parsing} (recovering structured fields such as author, title, venue/publisher, and date). In summary, providing robust and scalable citation processing of SSH literature is a major technical obstacle to overcome for building an SSH citation index \cite{colavizza_case_2022}, which constitutes the backdrop and ultimate objective of the work presented in this paper. 

A second, closely related issue is evaluation mismatch. Widely used benchmarks often center on clean, English end-of-section bibliographies, which can inflate perceived robustness and hide SSH-specific failure modes. Meanwhile, SSH-oriented datasets are rarer and often cover only parts of the pipeline (e.g., parsing from gold strings rather than end-to-end processing from PDFs). Consequently, it remains difficult to assess real-world progress—both for supervised pipelines and LLM-based systems—under the regimes most relevant to SSH digitization and citation indexing. 

In this paper, we address this gap with a benchmark designed around SSH-realistic conditions and operational constraints, and use it to assess whether contemporary LLMs are sufficiently reliable for reference extraction and parsing. We evaluate three complementary datasets: \cex, \excite, and \linkbook, capturing (i) controlled English journal references across disciplines, (ii) German/English documents with explicit citation-placement regimes (end-section, footnote-only, mixed), and (iii) humanities references with high stylistic and multilingual variability. We benchmark three tasks of increasing difficulty: reference extraction, reference parsing, and end-to-end document parsing. For comparability across heterogeneous systems, we use a schema-constrained evaluation: LLMs must produce strict JSON outputs, and \grobid predictions are mapped to the same target schema.

Our empirical focus is twofold. First, we provide a transparent, side-by-side comparison between a strong supervised baseline (\grobid) and several contemporary LLMs, including \deepseek, \mistralsmall, \gemma, and the \qwenfam family. Second, we study two strategies to boost performance in practice: (i) parameter-efficient LoRA adaptation for SSH-style parsing and (ii) document segmentation/pipelining strategies that trade global context for shorter, verifiable outputs. The resulting findings suggest a pragmatic hybrid direction: \grobid remains highly effective when its operating assumptions hold, but LLMs—especially when adapted—offer stronger robustness under distribution shift (multilinguality and footnote-heavy layouts). This naturally motivates \emph{routing} policies that allocate easy, in-distribution documents to \grobid and escalate high-risk cases to adapted LLMs.

\paragraph{Contributions.}
\begin{itemize}
    \item We introduce a unified SSH-focused benchmark spanning \cex, \excite, and \linkbook, and covering extraction, parsing, and end-to-end document parsing under a shared schema and evaluation protocol.
    \item We benchmark \grobid against a diverse set of LLMs (\deepseek, \mistralsmall, \gemma, \qwenfam), highlighting where LLMs are competitive or superior under SSH-typical distribution shifts.
    \item We quantify the impact of two practical levers—LoRA adaptation and segmentation/pipelining—showing consistent improvements in robustness and end-to-end utility.
    \item We synthesize these results into actionable guidance for hybrid deployment via routing, balancing throughput and reliability in large SSH collections.
\end{itemize}

\paragraph{Paper outline.}
Section~\ref{sec:related-work} situates our work in prior datasets and tool chains; Section~\ref{sec:datasets} describes the three datasets; Section~\ref{sec:setup} details tasks, systems, and evaluation; Section~\ref{sec:results} reports benchmarking results and error analyses; Sections~\ref{sec:lora} and~\ref{sec:segmentation} study LoRA and segmentation; and Section~\ref{sec:conclusion} concludes with implications for robust SSH citation processing.  Our code is available at: \href{https://github.com/odoma-ch/ssh-citation-index/tree/main/benchmarks}{Github}

\section{Related Work}
\label{sec:related-work}

Research on bibliographic reference extraction and parsing has long been shaped by two practical constraints: the limited availability of high-quality gold standards and a persistent mismatch between \emph{laboratory-style} citation data (clean end-section bibliographies in English) and the heterogeneity of real scholarly documents, particularly in the Social Sciences and Humanities (SSH). This section reviews the datasets most commonly used to develop and benchmark reference processing systems, alongside representative toolchains ranging from supervised pipelines to recent LLM-based approaches. This context motivates our benchmark design, which explicitly targets multilinguality, footnote-heavy layouts, and long-form documents---settings in which traditional systems are known to degrade.

Existing datasets vary widely in authenticity, scale, and coverage. \excite~\cite{hosseini_excite_2019} is notable for providing an SSH-oriented benchmark with a strong German component and an explicit mixture of end-section, footnote-only, and mixed citation regimes. Beyond reference strings, it includes intermediate artifacts such as page-layout exports and corrected XML, which enable evaluation of end-to-end robustness rather than string-level parsing alone. GEOcite~\cite{Birkeneder2022ExtractingLR} similarly contributes German-language scientific publications with extracted reference strings and field-level segmentation annotations, mainly supporting supervised training for metadata extraction. 
In addition, \cex~\cite{cioffi_data_2022,pagnotta_cex_2024} offers carefully aligned PDFs and Text Encoding Initiative (TEI) annotations compliant with the \grobid citation schema, yielding a controlled multi-disciplinary English benchmark where citation conventions differ across 27 SCImago subject categories. 
Complementary corpora leverage paired full-text markup: the XML Markup Evaluation Corpus~\cite{garnett2016xmlmarkupcorpus} and Open Research Europe (ORE)~\cite{ec2025openresearcheuropecorpus} provide Journal Article Tag Suite (JATS)-aligned representations that facilitate large-scale harvesting of reference lists with structured metadata. 
At the other extreme, GIANT~\cite{grennan_giant_2019} provides massive synthetic supervision by rendering Crossref-derived records into reference strings under many Citation Style Language (CSL) styles. While this scale is valuable for training data-hungry models and broad style coverage, the clean generation process underrepresents the noise, abbreviation practices, multilingual variation, and idiosyncratic source types that are common in SSH references. 
Finally, \linkbook~\cite{colavizza_annotated_2017} stands out as a humanities-first resource, offering tens of thousands of manually annotated references spanning footnotes and bibliographies, including primary sources and abbreviated forms, with fine-grained field markup; its stylistic diversity and historical variability make it particularly suitable for stress-testing parsers under SSH-typical conditions.

On the methods side, supervised pipelines remain the dominant baseline for reference extraction and parsing. \grobid~\cite{tkaczyk_cermine_2015,prasad_dataset_2019} is widely adopted for PDF-to-TEI conversion and frequently used as a comparative standard; recent benchmarking confirms its strong performance alongside systems such as AnyStyle on heterogeneous evaluation suites~\cite{backes_comparing_2024}. Nevertheless, multiple studies report brittleness under distribution shift, especially for multilingual content, complex page geometry, multi-column layouts, and footnote-embedded citations, where segmentation failures and missed boundaries become common~\cite{ghavimi2019exmatchercombiningfeaturesbased,adhikari2025comparativestudypdfparsing}. Motivated by these limitations, recent work has begun to apply LLMs directly to citation parsing. Sarin and Alperin~\cite{sarin_citation_2025} construct paired datasets from JATS sources and show that small open-weight LLMs can achieve strong field-level accuracy in few-shot settings, often surpassing previously reported results on similar tasks. However, their studies largely focus on English, DOI-rich journal references and do not systematically evaluate SSH-specific regimes (footnotes, long monographs, mixed-language corpora), nor do they provide transparent, side-by-side comparisons against strong \grobid configurations on the same inputs.

Overall, a consistent gap remains: evaluation conditions are frequently misaligned with SSH reality, where citations are embedded in footnotes, abbreviated, multilingual, and shaped by diverse historical conventions. Our benchmarking work is designed to address this gap by jointly testing extraction, parsing, and end-to-end behavior on \cex, \excite, and \linkbook, while retaining \grobid as a strong supervised baseline and extending comparison to contemporary LLMs and parameter-efficient adaptation.

\section{Datasets}
\label{sec:datasets}
For evaluation, we select three complementary datasets: the \textbf{\cex project Goldstandard}, the \textbf{EXparser Gold Standard} from the EXCITE project,  and \textbf{\linkbook}. They differ in domain, language distribution, and citation style (end-of-document bibliographies vs.\ footnotes), enabling robust benchmarking across document- and reference-level tasks. To avoid skewing the dataset towards simple, non-SSH citation styles, we deliberately exclude additional datasets with overlapping scope.
% (e.g. English language only, journal-style references without footnotes).  
% Figure~\ref{fig:dataset-stats} and 
Table~\ref{tab:datasets} summarize their main characteristics.

\begin{table}[!htbp]
\caption{Overview of benchmark datasets. \linkbook is released as reference-level records; document counts are not directly comparable.}
\label{tab:datasets}
\centering
\small
\begin{tabular}{l r r p{0.55\linewidth}}
\toprule
Dataset & \# Documents & \# References & Content \\
\midrule
\cex & 112 & 5,160 & English-language journal articles from 27 SCImago disciplines \\
\excite & 351 & 10,171 & Majority German; rich mix of end-section, footnote-only, and mixed citation layouts \\
\linkbook & N/A & 1,194 & Monographs and articles; multilingual references: IT (72.3\%), EN (18.3\%), DE (4.3\%), FR (3.9\%), ES (0.6\%), PT (0.5\%), NL (0.1\%) \\
\bottomrule
\end{tabular}
\end{table}

\subsection{\cex}
\label{sec:cex}
The \textbf{\cex project Goldstandard} (hereafter \cex) \cite{pagnotta_cex_2024} is a curated and corrected expansion of Cioffi \cite{cioffi_data_2022}, released as paired PDFs and TEI-encoded reference annotations compliant with the \grobid citation model. It contains 112 English papers (1,581 pages; 5,160 references) spanning 27 general SCImago disciplines, where reference formatting and citation conventions vary substantially across fields. We use the full set as a controlled benchmark to probe cross-domain generalization.

\subsection{\excite}
The \textbf{EXparser Gold Standard} (hereafter \excite) was created within the EXCITE project~\cite{hosseini_excite_2019}, a reference extraction and citation matching toolchain for German-language social-science publications,  including sociology and migration studies, political science, health and social policy, and economics. \excite is designed to support layout analysis, reference identification, and segmentation/metadata extraction. It comprises 351 SSH papers (8{,}041 pages; 10{,}171 references): 251 German documents (219 with end-section references, 12 with mixed end-section and footnotes, and 20 footnote-only) and 100 English documents with standard end-section bibliographies. Beyond layout variation, the footnote-heavy classes (2 and 3) introduce \textbf{abbreviated back-references} typical of German SSH writing: Ebd. (ebenda, equivalent to ibid.) and a.a.O. (am angegebenen Ort, equivalent to op. cit.). The gold standard retains these strings verbatim without resolving them to their canonical referents.

The gold standard provides multiple aligned representations per paper (PDF plus intermediate layout/annotation files), enabling evaluation of extraction robustness across layout and citation-placement conditions. We use the full gold standard in our experiments.

\subsection{\linkbook}

\linkbook~\cite{colavizza_annotated_2017} is a humanities-oriented citation mining project centered on the historiography of Venice. It provides 40k+ manually annotated references extracted from both bibliographies and footnotes, with fine-grained field segmentation and substantial variation in humanities citation style, including primary vs.\ secondary references and abbreviated forms. The corpus is released with predefined splits (train $\approx$35k, validation $\approx$2k, test $\approx$2k). For our benchmark, we evaluate on the full test split but restrict to \emph{primary} citations only. This yields 1{,}194 references, predominantly in Italian and English (72.3\% IT, 18.3\% EN), with smaller portions of German, French, Spanish, Portuguese, and Dutch. This multilingual and stylistically heterogeneous subset is well suited to stress-testing reference parsing in history- and humanities-style citation practices.

% \begin{figure}
% \centering
% \includegraphics[width=0.95\linewidth]{figs/dataset_statistics.png}
% \caption{Dataset comparison statistics across \cex, \excite, and \linkbook.
% Top: number of documents and references.
% Bottom: average references and pages per document.}
% \label{fig:dataset-stats}
% \end{figure}

\section{Setup}
\label{sec:setup}
\subsection{Tasks}

We evaluate all systems on three tasks of increasing complexity: \textbf{reference extraction}, \textbf{reference parsing}, and \textbf{end-to-end document parsing}. This setup separates span detection from structured metadata prediction and assesses their combined behavior in realistic pipelines.

\paragraph{Reference extraction.}
A document-level task: given a document (or segment), systems must detect and delimit all bibliographic references—whether in bibliographies or footnotes—and return them as plain-text spans without internal structure.

\paragraph{Reference parsing.}
A reference-level task with gold reference strings as input. The goal is to convert each string into a structured record (authors, title, year, venue/publisher, identifiers), isolating fine-grained semantic segmentation from extraction errors.

\paragraph{End-to-end document parsing.}
A pipeline task that jointly performs extraction and parsing from raw documents, producing structured reference records. Performance here reflects cumulative errors and thus approximates practical usability for citation indexing and linking.

\subsection{Systems and infrastructure}
We compare \grobid with several LLMs under a unified, schema-constrained setup.

\paragraph{\grobid baseline.}
We run \grobid with the \textbf{Deep Learning sequence labelling model} released by the CEX project~\cite{pagnotta_cex_2024-1}. The model is trained on citation data closely related to, but distinct from, our \cex gold standard (Section~\ref{sec:cex}). It thus provides a strong supervised baseline for English scientific articles and has a distributional advantage on \cex; we therefore emphasize cross-dataset performance on \excite and \linkbook to assess robustness on multilingual, heterogeneous SSH material. For the reference parsing task, we use \texttt{/api/processCitationList} endpoint on gold reference strings directly; for the end-to-end parsing, we use \texttt{/api/processFulltextDocument} endpoint on pdf ducuments.

\paragraph{LLM models.}
Our pool includes open-weight models that can be self-hosted and one model we access via API for convenience. \deepseek~\cite{deepseek-ai_deepseek-v3_2025} represents a very-large, state-of-the-art open-source LLM; although open weights permit self-hosting, we evaluate it through the public API in this study. Mistral-Small-3.2-24B-Instruct-2506 (hereafter \mistralsmall)~\cite{mistral_small_3_2025} and \gemma~\cite{gemma_team_gemma_2024} cover a practical range for local inference. The \qwenfam{} family (4B, 8B, 30B-A3B, 32B)~\cite{bai2025qwen3vltechnicalreport} enables controlled scaling comparisons within one architecture.We recognize that certain documents within our datasets are publicly accessible and may be included in the pretraining corpora of the assessed models. Nonetheless, there is no indication that any of the chosen LLMs were trained on the specific document-reference pairs utilized in our benchmark; the datasets also encompass closed-access publications that the models could not have accessed; furthermore, SSH citation processing is an under-represented domain in LLM training, thereby diminishing the likelihood of systematic bias.

\paragraph{Prompts, input representation, and decoding.}
All LLMs use a few-shot prompt with a fixed JSON schema (authors, full\_title, year, venue/publisher, identifiers, etc.)\footnote{The prompts we use can be found at: \url{https://github.com/odoma-ch/ssh-citation-index/tree/main/prompts}}. Prompts include brief instructions, an inline schema block, and 1--2 style-diverse examples. Outputs must be strict JSON (no prose). For document-level tasks (reference extraction and end-to-end parsing), we first convert each input PDF into a text-only Markdown representation (including footnote text when present) using the \texttt{marker} PDF-to-text pipeline,\footnote{\url{https://github.com/datalab-to/marker}} selected on the basis of an internal evaluation balancing extraction accuracy and processing speed, and provide the markdown text as the model input. GROBID, by contrast, ingests PDF files directly and performs its own layout analysis, so no intermediate conversion is applied. For reference parsing, the model input is the gold reference string and the PDF-to-Markdown step is bypassed. We use provider-recommended inference settings for each model (temperature and top\_p), and we do not impose additional output-length caps beyond the serving backend's defaults/limits. Generations are validated as JSON; if an output is invalid (e.g., malformed/truncated JSON or empty), we perform a single retry with identical settings, after which the instance is counted as a failure.

\paragraph{Infrastructure.}
Self-hosted models are served with \texttt{vLLM} on 2$\times$A100 NVIDIA GPUs; \deepseek{} is accessed via its public API; \grobid{} runs on a 24\,GB MIG slice. We report accuracy metrics alongside wall-clock runtime and failures (invalid JSON, empty outputs, truncation).

\subsection{Evaluation protocol}
\label{sec:eval}

We evaluate three tasks: reference extraction, reference parsing, and end-to-end document parsing. All tasks use the same soft one-to-one matching: each gold reference is paired with the most similar prediction using a \textbf{fuzzy Levenshtein similarity score} over the reference string and key fields. Prior to scoring, we case-fold and normalize whitespace/punctuation; names are canonicalized in an order-insensitive manner (e.g., ``Surname, First M.'' $\approx$ ``First M.\ Surname''). We do not apply hard similarity thresholds; \textbf{similarity directly determines partial credit in precision and recall.} We evaluate against \textbf{verbatim gold standard strings} in their original document order, without resolving abbreviated back-references (e.g., Ebd., a.a.O.). Downstream resolution is left to dedicated entity-linking components.

For \textbf{reference extraction} (string level), we match predicted reference strings to gold strings and report Precision, Recall, and F1. Extra duplicates count as false positives; unmatched gold references count as false negatives.

For \textbf{reference parsing} and \textbf{end-to-end document parsing} (field level), we score structured fields for each matched reference under the same rule. We report micro-F1 (pooled over records and fields), macro-F1 (averaged over records), and per-field precision/recall/F1 for key fields (Title, Authors, Year), with additional fields in detailed tables.

Invalid, empty, or ill-formed JSON yields no valid items (TP = 0) and contributes to errors through matching. We additionally track failure counts and runtime as operational indicators.

\section{Experiments and Results}
\label{sec:results}
\subsection{Off-the-shelf LLM baselines}

Our first experiment benchmarks off-the-shelf LLMs against the \grobid baseline in a single-call, few-shot setting (Section~\ref{sec:setup}), without task-specific fine-tuning. We evaluate reference extraction, reference parsing, and end-to-end parsing on \cex, \excite, and \linkbook. Table~\ref{tab:vanilla-result} reports the main results. 
% while Table~\ref{tab:field-f1-breakdown} in the appendix provides per-field F1 breakdowns for fields that hat occur in at least 50\% of the annotations within each dataset. 
We do not report a standalone \grobid score for reference extraction, as \grobid is designed as a pipeline and does not provide an extraction-only endpoint that isolates bibliography/footnote segmentation from downstream parsing.

\begin{table}[!htbp]
\caption{Benchmark results for reference extraction and parsing (single-call / few-shot). We report Precision (P), Recall (R), Micro F1, and Macro F1. Fail/Error counts include invalid/empty outputs for LLMs and pipeline failures for \grobid. Best Micro F1 per dataset within each task block is shown in bold; second-best is underlined.}
\label{tab:vanilla-result}
\centering
\scriptsize
\resizebox{\linewidth}{!}{%
\begin{tabular}{l l c c c c c c}
\toprule
Dataset & Model & P & R & Micro F1 & Macro F1 & \# Fail/Errors & Runtime (s) \\
\midrule
\multicolumn{8}{c}{\textbf{Reference Extraction}} \\
\midrule
\multirow{7}{*}{\cex}
  & \deepseek       & 0.9616 & 0.9196 & 0.9373 & -- & 0 & 566.74 \\
  & \mistralsmall   & 0.9906 & 0.9915 & \textbf{0.9910} & -- & 0 & 1315.98 \\
  & \gemma          & 0.7087 & 0.6153 & 0.6413 & -- & 0 & 1889.41 \\
  & \qwen{32B}      & 0.9681 & 0.9703 & \underline{0.9689} & -- & 0 & 1826.02 \\
  & \qwen{30B-A3B}  & 0.9609 & 0.9614 & 0.9611 & -- & 0 & 656.30 \\
  & \qwen{8B}       & 0.9591 & 0.9559 & 0.9572 & -- & 0 & 624.24 \\
  & \qwen{4B}       & 0.9652 & 0.9632 & 0.9629 & -- & 0 & 1625.35 \\
\cmidrule(l){1-8}
\multirow{7}{*}{\excite}
  & \deepseek       & 0.9472 & 0.9328 & \underline{0.9308} & -- & 2 & 846.02 \\
  & \mistralsmall   & 0.8956 & 0.9261 & 0.8984 & -- & 3 & 2965.49 \\
  & \gemma          & 0.7387 & 0.7573 & 0.7348 & -- & 2 & 2202.33 \\
  & \qwen{32B}      & 0.9446 & 0.9421 & \textbf{0.9367} & -- & 0 & 1804.71 \\
  & \qwen{30B-A3B}  & 0.9346 & 0.9344 & 0.9263 & -- & 0 & 1486.20 \\
  & \qwen{8B}       & 0.9380 & 0.9361 & 0.9285 & -- & 2 & 2355.55 \\
  & \qwen{4B}       & 0.9256 & 0.9298 & 0.9184 & -- & 5 & 2242.79 \\

\midrule
\multicolumn{8}{c}{\textbf{Reference Parsing}} \\
\midrule
\multirow{8}{*}{\cex}
  & \deepseek       & 0.3966 & 0.3805 & 0.3863 & 0.3779 & 0 & 1344.44 \\
  & \mistralsmall   & 0.7422 & 0.7401 & 0.7401 & 0.7350 & 0 & 1569.56 \\
  & \gemma          & 0.7575 & 0.7092 & 0.7233 & 0.7046 & 2 & 4349.06 \\
  & \qwen{32B}      & 0.7502 & 0.7461 & \underline{0.7475} & 0.7419 & 0 & 1989.12 \\
  & \qwen{30B-A3B}  & 0.7336 & 0.7351 & 0.7337 & 0.7271 & 0 & 693.55 \\
  & \qwen{8B}       & 0.7386 & 0.7229 & 0.7293 & 0.7197 & 0 & 762.24 \\
  & \qwen{4B}       & 0.7269 & 0.7115 & 0.7173 & 0.7101 & 1 & 3870.84 \\
  & \grobid         & 0.9120 & 0.8081 & \textbf{0.8560} & 0.8457 & 4 & 1362 \\
\cmidrule(l){1-8}
\multirow{8}{*}{\excite}
  & \deepseek       & 0.8753 & 0.7838 & 0.8233 & 0.8054 & 12 & 1945.15 \\
  & \mistralsmall   & 0.8997 & 0.8342 & \textbf{0.8623} & 0.8411 & 2 & 1409.64 \\
  & \gemma          & 0.8877 & 0.7648 & 0.8167 & 0.8029 & 5 & 3334.53 \\
  & \qwen{32B}      & 0.8956 & 0.8237 & \underline{0.8550} & 0.8354 & 1 & 2629.65 \\
  & \qwen{30B-A3B}  & 0.8659 & 0.8295 & 0.8448 & 0.8255 & 6 & 1943.02 \\
  & \qwen{8B}       & 0.8870 & 0.8131 & 0.8462 & 0.8293 & 5 & 1886.02 \\
  & \qwen{4B}       & 0.8562 & 0.7818 & 0.8151 & 0.7986 & 4 & 1832.85 \\
  & \grobid         & 0.7688 & 0.6310 & 0.6918 & 0.6853 & 0 & 2083 \\
\cmidrule(l){1-8}
\multirow{8}{*}{\linkbook}
  & \deepseek       & 0.8714 & 0.7476 & \textbf{0.8047} & 0.7999 & 4 & 305.61 \\
  & \mistralsmall   & 0.6682 & 0.8932 & 0.7645 & 0.7692 & 0 & 227.96 \\
  & \gemma          & 0.6853 & 0.5200 & 0.5907 & 0.5493 & 0 & 405.74 \\
  & \qwen{32B}      & 0.6868 & 0.8956 & \underline{0.7774} & 0.7716 & 0 & 403 \\
  & \qwen{30B-A3B}  & 0.6336 & 0.8689 & 0.7328 & 0.7237 & 1 & 349.00 \\
  & \qwen{8B}       & 0.6663 & 0.8577 & 0.7500 & 0.7418 & 0 & 358.01 \\
  & \qwen{4B}       & 0.6382 & 0.6587 & 0.6483 & 0.5968 & 1 & 1209.67 \\
  & \grobid         & 0.6615 & 0.7138 & 0.6866 & 0.6517 & 0 & 251 \\

\midrule
\multicolumn{8}{c}{\textbf{End-to-end Parsing}} \\
\midrule
\multirow{8}{*}{\cex}
  & \deepseek       & 0.4722 & 0.4421 & 0.4553 & 0.4439 & 3 & 2392.30 \\
  & \mistralsmall   & 0.6356 & 0.6044 & 0.6076 & 0.5922 & 0 & 3645.20 \\
  & \gemma          & 0.6452 & 0.3606 & 0.4265 & 0.3577 & 1 & 1387.60 \\
  & \qwen{32B}      & 0.7325 & 0.7241 & \underline{0.7269} & 0.7182 & 0 & 2048.62 \\
  & \qwen{30B-A3B}  & 0.7032 & 0.6972 & 0.6984 & 0.6868 & 6 & 4472.04 \\
  & \qwen{8B}       & 0.6867 & 0.6769 & 0.6801 & 0.6710 & 1 & 3108.10 \\
  & \qwen{4B}       & 0.6606 & 0.6472 & 0.6488 & 0.6368 & 7 & 4497.63 \\
  & \grobid         & 0.8862 & 0.7850 & \textbf{0.8314} & 0.8090 & 0 & 806.28 \\
\cmidrule(l){1-8}
\multirow{8}{*}{\excite}
  & \deepseek       & 0.8408 & 0.6925 & \underline{0.7518} & 0.7195 & 10 & 2158.85 \\
  & \mistralsmall   & 0.6149 & 0.5322 & 0.5594 & 0.5339 & 3 & 4746.59 \\
  & \gemma          & 0.7034 & 0.5097 & 0.5673 & 0.5237 & 3 & 5113.73 \\
  & \qwen{32B}      & 0.8189 & 0.7051 & \textbf{0.7524} & 0.7278 & 4 & 5117.88 \\
  & \qwen{30B-A3B}  & 0.7703 & 0.6739 & 0.7091 & 0.6767 & 7 & 3923.41 \\
  & \qwen{8B}       & 0.8079 & 0.6975 & 0.7432 & 0.7145 & 10 & 3367.76 \\
  & \qwen{4B}       & 0.7855 & 0.6822 & 0.7234 & 0.6999 & 34 & 5126.05 \\
  & \grobid         & 0.6163 & 0.5117 & 0.5477 & 0.5030 & 41 & 1813.20 \\

\bottomrule
\end{tabular}
}%
\end{table}

\paragraph{Task-wise performance.}
Across tasks, the LLMs and \grobid exhibit distinct performance regimes. For \textbf{reference extraction}, the larger LLMs largely saturate the problem on both \cex and \excite: once the model has sufficient capacity, span identification is highly reliable and invalid outputs are rare. Differences between top models are small, suggesting limited sensitivity to modeling choices.

For \textbf{reference parsing}, outcomes depend more on domain match and supervision. On \cex, \grobid leads, consistent with its in-domain training advantage and its role as a strong supervised baseline. On \excite and \linkbook, however, several LLMs match or surpass \grobid, indicating better cross-style generalization and greater robustness under distribution shift.

\textbf{End-to-end} results mirror this trade-off. \grobid performs best on \cex where its advantage compounds across stages, while on \excite the strongest LLMs overtake the \grobid by retaining more stable parsing under non-standard layouts and multilingual references.

\paragraph{Scaling effects.}
After a moderate capacity threshold, model size has limited impact on \textbf{extraction}, but plays a much larger role in \textbf{parsing} and \textbf{end-to-end} parsing, where systems must maintain structured, well-formed outputs over long and noisy inputs. This pattern holds both \emph{within} and \emph{across} model families: smaller variants (e.g., \qwen{4B}) exhibit more failures and less stable behavior, while larger models are generally more robust and accurate.

Within \qwenfam, moving from \qwen{4B} to \qwen{8B} yields a clear jump in reliability, and the 30B-A3B and 32B models often approach or exceed the frontier-level performance of \deepseek, matching or outperforming \grobid on heterogeneous SSH datasets. At the same time, these gains are not strictly monotonic with parameter count: \mistralsmall is frequently among the strongest models on \excite parsing and can outperform \qwen{32B} in some settings, while \qwen{4B} occasionally surpasses larger models such as \gemma and \mistralsmall on specific datasets or fields. In practice, scaling laws still provide a useful aggregate trend, but architectural design, instruction tuning, and model ``freshness'' can be at least as important as raw size when choosing a model for structured citation prediction.

\paragraph{SSH factors}
The \excite breakdown in Table~\ref{tab:excite-breakdown} highlights how SSH-specific characteristics, i.e., language and citation placement shape performance. 
For LLMs, English vs.\ German bibliographies (Class~1) differ only slightly, whereas \grobid shows a language clearer gap under conventional layouts.

In German, shifting from bibliography-style references to \textbf{footnote-only} and \textbf{mixed} regimes causes a large drop for all systems, but \grobid deteriorates most sharply. LLMs are consistently more robust in footnote-heavy settings, with \deepseek exhibiting the smallest placement-induced gaps, consistent with stronger generalization at higher capacity. Mixed regimes remain difficult across the board, likely because they combine segmentation ambiguity with cross-location consistency constraints.

Overall, LLMs are better suited to heterogeneous SSH citation practices than \grobid, but footnote-driven layouts remain the dominant failure mode for all systems.

\begin{table}[htb]
\caption{Performance breakdown on the \excite dataset by reference class and language.
We report Precision (P), Recall (R), and F1 for reference extraction, reference parsing,
and end-to-end parsing. Classes correspond to references appearing
(1) exclusively in end-of-document bibliographies,
(2) exclusively in footnotes, and
(3) jointly in footnotes and end-of-document sections.}
\label{tab:excite-breakdown}
\centering
\small
\resizebox{\linewidth}{!}{%
\begin{tabular}{llccccc}
\toprule
\textbf{Task} & \textbf{Class / Lang} &
\textbf{\deepseek} &
\textbf{\mistralsmall} &
\textbf{\gemma} &
\textbf{\qwen{32B}} &
\textbf{\grobid} \\
\midrule
\multirow{4}{*}{\textbf{Extraction}}
 & Class 1 (EN) & 0.97 / 0.98 / 0.98 & 0.95 / 0.98 / 0.96 & 0.78 / 0.80 / 0.78 & 0.98 / 0.98 / 0.98 & -- \\
 & Class 1 (DE) & 0.94 / 0.95 / 0.94 & 0.88 / 0.94 / 0.90 & 0.73 / 0.78 / 0.74 & 0.95 / 0.96 / 0.95 & -- \\
 & Class 2 (DE) & 0.83 / 0.69 / 0.73 & 0.78 / 0.73 / 0.73 & 0.65 / 0.49 / 0.54 & 0.71 / 0.74 / 0.71 & -- \\
 & Class 3 (DE) & 0.81 / 0.46 / 0.57 & 0.80 / 0.46 / 0.56 & 0.56 / 0.35 / 0.41 & 0.70 / 0.36 / 0.46 & -- \\
\midrule
\multirow{4}{*}{\textbf{Parsing}}
 & Class 1 (EN) & 0.88 / 0.82 / 0.85 & 0.92 / 0.89 / 0.90 & 0.91 / 0.82 / 0.86 & 0.92 / 0.89 / 0.90 & 0.83 / 0.69 / 0.75 \\
 & Class 1 (DE) & 0.89 / 0.79 / 0.83 & 0.90 / 0.83 / 0.86 & 0.89 / 0.76 / 0.82 & 0.90 / 0.81 / 0.85 & 0.77 / 0.63 / 0.69 \\
 & Class 2 (DE) & 0.84 / 0.68 / 0.74 & 0.84 / 0.68 / 0.74 & 0.81 / 0.65 / 0.72 & 0.82 / 0.75 / 0.78 & 0.48 / 0.41 / 0.44 \\
 & Class 3 (DE) & 0.70 / 0.50 / 0.57 & 0.89 / 0.62 / 0.71 & 0.75 / 0.48 / 0.56 & 0.79 / 0.57 / 0.65 & 0.61 / 0.51 / 0.56 \\
\midrule
\multirow{4}{*}{\textbf{End-to-end}}
 & Class 1 (EN) & 0.89 / 0.78 / 0.83 & 0.75 / 0.67 / 0.70 & 0.77 / 0.59 / 0.65 & 0.89 / 0.79 / 0.83 & 0.75 / 0.65 / 0.69 \\
 & Class 1 (DE) & 0.83 / 0.68 / 0.74 & 0.58 / 0.51 / 0.53 & 0.69 / 0.52 / 0.57 & 0.80 / 0.69 / 0.74 & 0.59 / 0.49 / 0.52 \\
 & Class 2 (DE) & 0.68 / 0.48 / 0.55 & 0.47 / 0.32 / 0.37 & 0.49 / 0.16 / 0.23 & 0.67 / 0.56 / 0.60 & 0.22 / 0.14 / 0.16 \\
 & Class 3 (DE) & 0.78 / 0.42 / 0.51 & 0.30 / 0.15 / 0.19 & 0.59 / 0.25 / 0.30 & 0.70 / 0.32 / 0.43 & 0.61 / 0.40 / 0.45 \\
\bottomrule
\end{tabular}}
\end{table}

\begin{figure}[t]
  \centering
  % \begin{minipage}[t]{0.8\linewidth}
  %   \centering
    \includegraphics[width=0.49\linewidth]{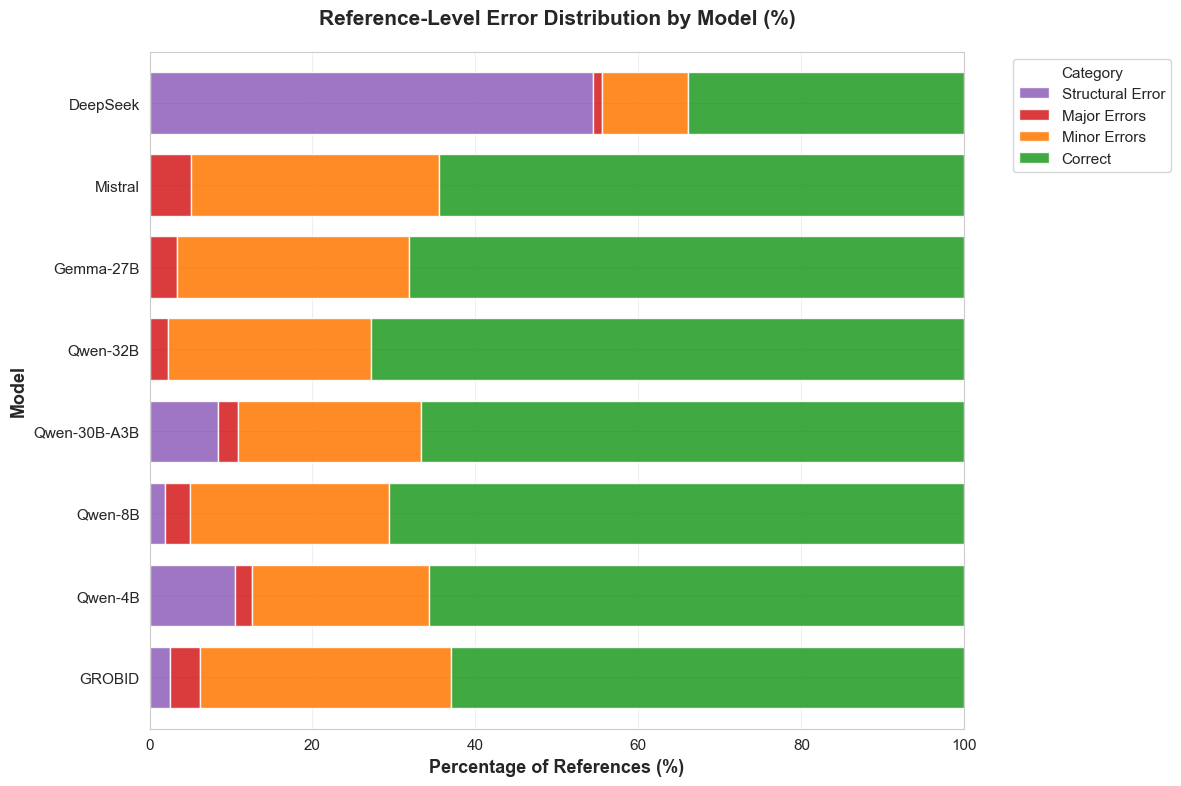}
  % \end{minipage}\hfill
  % \begin{minipage}[t]{0.7\linewidth}
  %   \centering
    \includegraphics[width=0.47\linewidth]{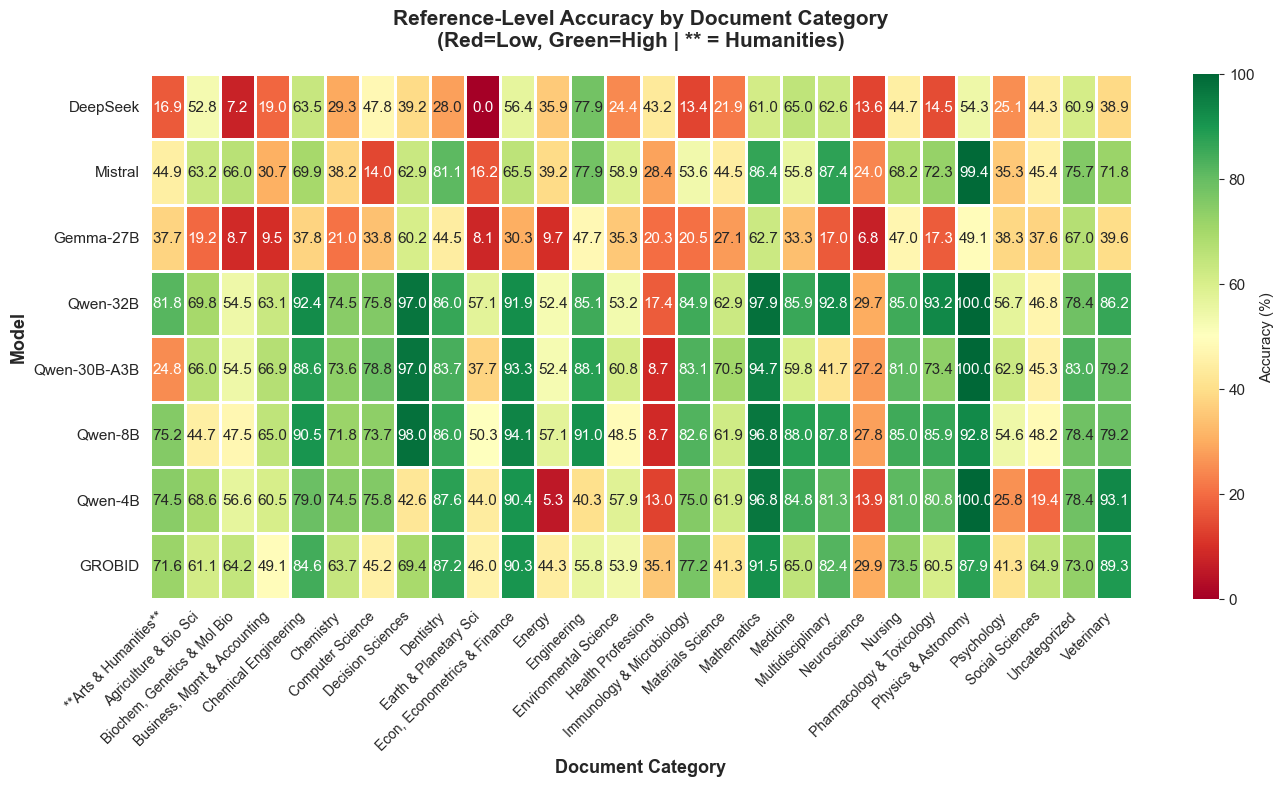}
  % \end{minipage}
\caption{Reference-level error analysis on CEX dataset and end-to-end document parsing task. 
\textbf{(a)} Overall error distribution by model/system, showing percentage of references in each category: structural errors, major errors, minor errors, and correct matches. 
\textbf{(b)} Reference-level accuracy (\%) by document category across models; color indicates performance (red = low, green = high). Categories marked with \texttt{**} denote humanities subjects.}
  \label{fig:ref_error_analysis}
\end{figure}

\paragraph{Error analysis. }
\label{sec:error-analysis}
Figure~\ref{fig:ref_error_analysis} provides a reference-level view of failure modes on \cex, complementing the F1-based evaluation in the previous section. For this analysis, we first reconstruct a canonical reference string from each predicted JSON record (by concatenating key fields in a fixed order) and compare it to the gold reference string using the same fuzzy-matching procedure as in Section~\ref{sec:setup}. The per-field similarities are then collapsed into four coarse categories:
\emph{Correct} if all fields reach a similarity of at least 0.95; \emph{Minor error} if at least one field falls into the $[0.60,0.95)$ band while none drops below 0.60; \emph{Major error} if any key field is below 0.60 but the JSON is structurally valid; and \emph{Structural error} if the output cannot be parsed into a reference at all (e.g., malformed or truncated JSON). 
% Table~\ref{tab:error-examples} in the appendix displays these categories with concrete examples of the four error types drawn from the \cex end-to-end parsing task.

Figure~\ref{fig:ref_error_analysis}\textbf{(a)} shows the resulting distribution over these categories, aggregated across all references. Most open-weight models and \grobid produce a large majority of \emph{Correct} references, with residual mistakes dominated by \emph{Minor} rather than \emph{Major} errors, i.e., by field omissions or local formatting issues instead of completely mismatched records. \deepseek remains an outlier with a comparatively high fraction of \emph{Structural} errors, which we attribute primarily to output-length limits in the API: for documents with long bibliographies, generations are frequently cut mid-output, yielding incomplete JSON that cannot be evaluated. At the same time, the stacked bars make a scaling effect visible within the \qwenfam line: the smallest variant \qwen{4B} and \qwen{30B-A3B} show noticeably more \emph{Structural} failures than its larger counterparts, whereas structural errors almost disappear for \qwen{32B}. This pattern supports the main benchmark finding that below a certain capacity threshold models become unstable on long, schema-constrained outputs, even when their token-level accuracy is competitive.

Figure~\ref{fig:ref_error_analysis}\textbf{(b)} breaks down the share of \emph{Correct} references (our strict $\geq\!0.95$ threshold on all fields) by document category. A consistent trend is that all systems achieve higher accuracy on STEM-oriented categories than on SSH/humanities-marked fields (indicated with \texttt{**}), suggesting increased sensitivity to stylistic variation, multilinguality, and heterogeneous citation conventions in the latter. Two categories stand out as difficult across models: \emph{Health Professions} and \emph{Neuroscience}, which show pronounced drops in accuracy. For \emph{Neuroscience}, this correlates with unusually long reference lists in \cex (roughly $75$ references per paper, versus an overall average of $\sim 41$), increasing the risk of truncation and accumulated formatting errors; both categories also contain highly specialised terminology and venue-specific citation patterns that appear less well covered by general instruction-tuned LLMs, leading to field confusions (e.g., identifiers mislabelled as page ranges). These category-level differences should nonetheless be interpreted with caution, as each \cex subject area is represented by only four documents, and some effects may therefore reflect sampling noise rather than stable disciplinary trends.

\subsection{LoRA Fine-tuning}
\label{sec:lora}

We investigate parameter-efficient fine-tuning, more specifically Low-Rank Adaptation (LoRA)\cite{hu2021loralowrankadaptationlarge}, to adapt locally deployable LLMs to SSH-typical reference parsing, using two open-weight backbones: \mistralsmall and \qwen{8B}. We train LoRA adapters for \textbf{reference parsing}  on a supervised dataset \footnote{Training dataset available at \url{https://huggingface.co/datasets/odoma/reference-parsing-finetuning}.}  we construct from the same three datasets used in our benchmark (\cex, \excite, \linkbook). Since \cex\ and \excite\ provide no dedicated training corpora, we sample from their gold references and enforce document-level exclusion from evaluation to prevent leakage. To balance scale across sources, we subsample roughly 10\% of the \linkbook training/validation material and retain only high-quality primary citations. To better reflect production inputs, we also create ``grouped'' instances by concatenating references that co-occur in the same gold reference block, apply our conversational templates to these blocks, and supervise the corresponding JSON lists, yielding 1{,}832 training conversations in total.

We train with Axolotl\cite{axolotl_maintainers_and_contributors_axolotl_2023} on two A100 GPUs, freezing base weights and selecting hyper-parameters via pilot sweeps. Best settings are: \mistralsmall\footnote{This model is available at: \url{https://huggingface.co/odoma/Mistral-Small-3.2-24B-LoRA-ref-parsing}} ($r{=}32$, $\alpha{=}64$, lr $5{\times}10^{-5}$, 2 epochs) and \qwen{8B} \footnote{This model is available at: \url{https://huggingface.co/odoma/Qwen3-VL-8B-LoRA-ref-parsing}} ($r{=}64$, $\alpha{=}128$, lr $5{\times}10^{-5}$, 3 epochs). To separate parameter adaptation from prompt effects, we evaluate three inference variants per backbone and dataset (Table~\ref{tab:lora-result}): (i) base model with our detailed few-shot prompt; (ii) LoRA model with the same detailed few-shot prompt; and (iii) LoRA model with the short, zero-shot \emph{training-style} prompt used during fine-tuning.

\begin{table}[thb]
\caption{
Reference parsing results for LoRA fine-tuning (\grobid as baseline). We compare (i) base models with few-shot prompting, (ii) LoRA-finetuned models with the same prompt, and (iii) LoRA-finetuned models with the zero-shot training-style prompt. Best Micro F1 per dataset and backbone in bold. (ties included).
}
\label{tab:lora-result}
\centering
\scriptsize
\resizebox{\linewidth}{!}{%
\begin{tabular}{l l c c c c c c}
\toprule
Dataset & Model variant & P & R & Micro F1 & Macro F1 & \# Fail & Runtime (s) \\
\midrule
\cex 
& Mistral-24B (base, few-shot) & 0.7422 & 0.7401 & 0.7401 & 0.7350 & 0 & 1569.6 \\
& Mistral-24B (LoRA, few-shot) & 0.7470 & 0.7420 & \textbf{0.7450} & 0.7426 & 0 & 1498.0 \\
& Mistral-24B (LoRA, training-style) & 0.7486 & 0.7420 & \textbf{0.7450} & 0.7548 & 0 & 1311.0 \\
& Qwen3-VL-8B (base, few-shot) & 0.7386 & 0.7229 & 0.7293 & 0.7197 & 0 & 762.2 \\
& Qwen3-VL-8B (LoRA, few-shot) & 0.7507 & 0.7398 & \textbf{0.7445} & 0.7475 & 0 & 841.6 \\
& Qwen3-VL-8B (LoRA, training-style) & 0.7480 & 0.7265 & 0.7358 & 0.7212 & 0 & 740.6 \\
& \grobid & 0.9120 & 0.8081 & 0.8560 & 0.8457 & 4 & 1362 \\
\midrule
\excite
& Mistral-24B (base, few-shot) & 0.8997 & 0.8342 & 0.8623 & 0.8411 & 2 & 1409.6 \\
& Mistral-24B (LoRA, few-shot) & 0.9310 & 0.8470 & 0.8840 & 0.8692 & 2 & 1543.0 \\
& Mistral-24B (LoRA, training-style) & 0.9340 & 0.8540 & \textbf{0.8904} & 0.8764 & 2 & 1523.0 \\
& Qwen3-VL-8B (base, few-shot) & 0.8870 & 0.8131 & 0.8462 & 0.8293 & 5 & 1886.0 \\
& Qwen3-VL-8B (LoRA, few-shot) & 0.9218 & 0.8039 & 0.8540 & 0.8321 & 0 & 1557.1 \\
& Qwen3-VL-8B (LoRA, training-style) & 0.9237 & 0.8192 & \textbf{0.8613} & 0.8373 & 0 & 1499.0 \\
& \grobid & 0.7688 & 0.6310 & 0.6918 & 0.6853 & 0 & 2083 \\
\midrule
\linkbook
& Mistral-24B (base, few-shot) & 0.6549 & 0.8822 & 0.7517 & 0.7473 & 0 & 131.7 \\
& Mistral-24B (LoRA, few-shot) & 0.7664 & 0.8879 & 0.8227 & 0.8177 & 0 & 122.0 \\
& Mistral-24B (LoRA, training-style) & 0.7768 & 0.8825 & \textbf{0.8293} & 0.8214 & 0 & 167.1 \\
& Qwen3-VL-8B (base, few-shot) & 0.6629 & 0.8443 & 0.7427 & 0.7284 & 0 & 62.0 \\
& Qwen3-VL-8B (LoRA, few-shot) & 0.7130 & 0.8936 & \textbf{0.8211} & 0.8296 & 0 & 68.5 \\
& Qwen3-VL-8B (LoRA, training-style) & 0.7490 & 0.8847 & 0.8112 & 0.8081 & 0 & 81.6 \\
& \grobid         & 0.6615 & 0.7138 & 0.6866 & 0.6517 & 0 & 251 \\
\bottomrule
\end{tabular}}
\end{table}

\paragraph{Results} LoRA consistently improves over the base models across all datasets, with the largest gains on SSH-heavy benchmarks. On \cex, improvements are modest but stable (Mistral: +0.7\% relative micro-F1; Qwen: +2.1\%). On \excite, both models improve more substantially (Mistral: +2.5\% with few-shot, +3.3\% with training-style prompting; Qwen: +0.9\% and +1.8\%). The strongest effect is on \linkbook, where both backbones gain $\approx$+10\% relative micro-F1 (Mistral: +9.4\% and +10.3\%; Qwen: +10.6\%). Prompt--training alignment yields additional but non-uniform benefits: it helps Mistral across all datasets and Qwen on \excite, while Qwen on \cex\ and \linkbook still prefers the richer few-shot prompt, suggesting that detailed instructions remain useful for highly variable humanities references even after adaptation. Operationally, failures remain zero in Table~\ref{tab:lora-result}, indicating preserved JSON validity; runtime differences mainly track prompt length, with training-style prompting generally reducing latency for Mistral.

Overall, lightweight adaptation with a small SSH supervision set (1{,}832 instances) substantially improves local reference parsing---especially under multilingual and style-rich distributions---and in the best cases enables more compact, schema-aligned prompting.

\subsection{Segmentation strategies}
\label{sec:segmentation}
\begin{figure}
    \centering
    \includegraphics[width=0.9\linewidth]{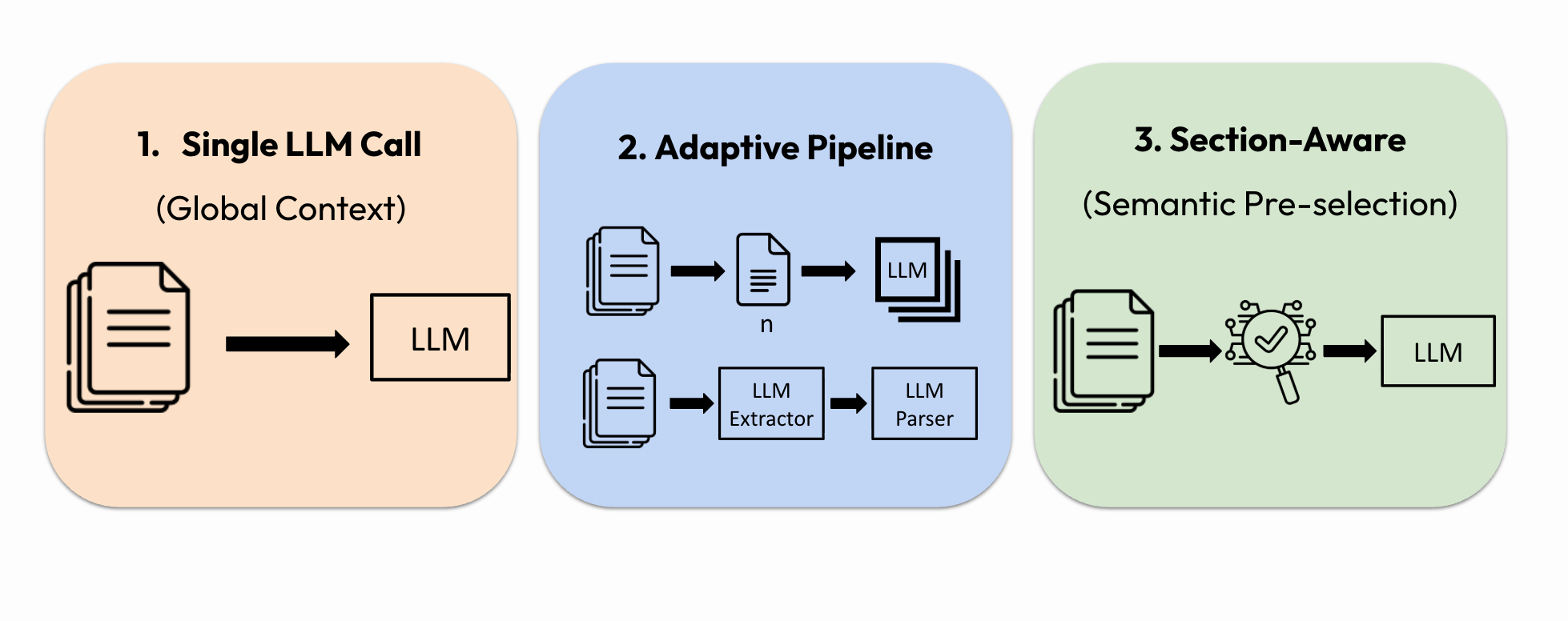}
    \caption{Segmentation strategies for reference extraction and end-to-end parsing.}
    \label{fig:segmentation}
\end{figure}

Segmentation controls how much text the model sees per request and therefore drives the trade-off between accuracy, robustness, and runtime. Rather than tuning segmentation per document, we treat it as an experimental factor and apply each strategy uniformly on \cex{} and \excite{} to isolate its effects under SSH-typical conditions (long documents, multilingual passages, and references in both end sections and footnotes). We compare three variants (Figure~\ref{fig:segmentation}):

\begin{itemize}
    \item \textbf{Single-call (global context)} sends the full document (or one long slice) in a single request when it fits the context window, preserving global coherence but becoming brittle as inputs and JSON outputs grow long.
    \item \textbf{Adaptive pipeline (local context)} reduces task complexity by breaking a single request into multiple smaller ones: for extraction, we apply \textbf{per-page segmentation}; for end-to-end parsing, we decompose the task into a sequential \textbf{extract $\rightarrow$ parse} pipeline. This keeps structured outputs small and localizes failures.
    \item \textbf{Semantic pre-selection (section-aware)} retrieves likely reference-bearing regions and only sends those chunks to the model, reducing token usage; we use \texttt{intfloat/multilingual-e5-large-instruct} for semantic retrieval.
\end{itemize}

Across variants, we keep the prompt and JSON schema fixed, enforce strict JSON validation, and log wall-clock runtime and invalid/empty outputs as failures.

\paragraph{Results}
Table~\ref{tab:segmentation-summary} summarizes results for \mistralsmall{} on \cex{} and \excite{}. For \textbf{reference extraction}, performance is high and differences are small: single-call is consistently best, the per-page adaptive setting is slightly weaker and can be slower on long documents, and semantic pre-selection largely preserves single-call accuracy while reducing runtime (notably on \excite{}).

For \textbf{end-to-end parsing}, segmentation has a much larger impact. Single-call is the least reliable, while the two-step extract$\rightarrow$parse pipeline performs best on both datasets by keeping structured outputs short and failures local. Semantic pre-selection is fastest and improves over single-call, but can miss scattered or stylistically atypical reference regions (e.g., footnote-heavy pages), limiting recall.

Overall, segmentation is secondary for extraction but critical for end-to-end parsing: routing the model through short, verifiable units (or selectively retrieving reference-bearing regions) is key to robust structured output on SSH documents.

\begin{table}[htb]
\caption{Segmentation comparison with \mistralsmall{} on \cex{} and \excite{}. Method~2 uses per-page segmentation for extraction and a two-step extract$\rightarrow$ parse pipeline for end-to-end parsing. Best values bolded.}
\label{tab:segmentation-summary}
\centering
\scriptsize
\setlength{\tabcolsep}{4pt}
\resizebox{\linewidth}{!}{%
\begin{tabular}{l l c c c c c c c c c c}
\toprule
\multirow{2}{*}{\textbf{Dataset}} &
\multirow{2}{*}{\textbf{Method}} &
\multicolumn{5}{c}{\textbf{Extraction}} &
\multicolumn{5}{c}{\textbf{End-to-end parsing}} \\
\cmidrule(lr){3-7}\cmidrule(lr){8-12}
& & \textbf{P} & \textbf{R} & \textbf{F1} & \textbf{AvgSim} & \textbf{Runtime (s)}
  & \textbf{P} & \textbf{R} & \textbf{F1} & \textbf{M-F1} & \textbf{Runtime (s)} \\
\midrule
\multirow{3}{*}{\cex{}} 
& 1~~Single-call & \textbf{0.9906} & \textbf{0.9915} & \textbf{0.9910} & \textbf{0.9973} & 1315.98 & 0.6356 & 0.6044 & 0.6076 & 0.5922 & 3645.20 \\
& 2~~Adaptive (per-page / two-step) & 0.9339 & 0.8958 & 0.9021 & 0.9527 & 2981.21 & \textbf{0.7467} & \textbf{0.7321} & \textbf{0.7364} & \textbf{0.7256} & 3059.09 \\
& 3~~Semantic pre-selection & 0.9469 & 0.9442 & 0.9446 & 0.9641 & \textbf{1162.16} & 0.7025 & 0.6550 & 0.6724 & 0.6506 & \textbf{1672.20} \\
\midrule
\multirow{3}{*}{\excite{}} 
& 1~~Single-call & 0.8956 & \textbf{0.9261} & \textbf{0.8984} & \textbf{0.9616} & 4576.27 & 0.6149 & 0.5322 & 0.5594 & 0.5339 & 4746.59 \\
& 2~~Adaptive (per-page / two-step) & \textbf{0.8967} & 0.9072 & 0.8894 & 0.9553 & 4377.95 & \textbf{0.8522} & \textbf{0.7913} & \textbf{0.8069} & \textbf{0.7700} & 4960.18 \\
& 3~~Semantic pre-selection & 0.8576 & 0.8582 & 0.8444 & 0.9170 & \textbf{2904.78} & 0.7276 & 0.6193 & 0.6565 & 0.6251 & \textbf{2702.19} \\
\bottomrule
\end{tabular}}
\end{table}

\section{Conclusion and Future Work}
\label{sec:conclusion}

This paper benchmarked bibliographic reference processing under SSH-realistic conditions, using \cex, \excite, and \linkbook across three tasks: reference extraction, reference parsing, and end-to-end document parsing. We compared a strong supervised pipeline baseline (\grobid) against contemporary LLMs under a unified, schema-constrained setup, and further studied practical deployment levers including LoRA adaptation and document segmentation strategies. 

Across datasets, extraction is comparatively stable: beyond a moderate capability threshold, most models reach similarly high accuracy and differences are driven more by throughput and failure rates than by peak scores. In contrast, parsing and end-to-end parsing remain the key bottlenecks, as they require consistent structured outputs and long-range robustness under noisy layouts. Here, the results reveal a clear trade-off: \grobid is typically the fastest option and excels on documents close to its training assumptions in the end-to-end document parsing task, but it degrades more strongly under distribution shift, especially for languages beyond English and for references embedded in footnotes or mixed citation regimes, where LLMs, especially after LoRA finetuning, are markedly more robust and often yield higher utility.

These findings suggest \textbf{LLM-routing} \cite{cooper_rethinking_2025} as a practical way forward to combine speed and robustness at scale. Data produced during evaluation can be used to train a small LLM to function as a router capable of dispatching the input document to the best-performing LLM or system, based on evaluation results (triplets of input document, model name, and F1-score). This approach would allow us to send to \grobid documents that match its operating envelope (e.g., English, well-structured PDFs with end-section bibliographies), while redirecting documents with known risk factors (multilinguality, dense footnotes, complex layouts) to task-adapted LLMs.  

Moreover, the stronger generalization of LLMs compared to supervised systems such as \grobid makes them particularly promising for SSH corpora, where complex layouts and multilinguality are the norm rather than the exception. This is especially relevant for highly multilingual collections, where a natural direction is to \emph{decompose} adaptation into modular components: (i) language-specific continuous pretraining for low-represented language (e.g., Latin, Polish), and (ii) task-specific LoRA adapters that capture the structured-output requirements of reference parsing. Composing these adapters separates linguistic coverage from task structure, enabling targeted updates without retraining the full system. Looking ahead, while our LoRA experiments target reference parsing, extending parameter-efficient adaptation to end-to-end document parsing remains an important next step. We will investigate LoRA for the full pipeline—layout analysis, reference extraction, and structured parsing—to improve overall reliability and reduce brittleness under SSH-specific shifts.

\begin{acknowledgments}
This work  was carried out in the context of the GRAPHIA project and received funding from the European Union (grant ID: 101188018, HORIZON-INFRA-2024-TECH-01 call) and from the Swiss State Secretariat for Education, Research and Innovation (SERI). Views and opinions expressed are however those of the author(s) only and do not necessarily reflect those of the European Union or the Agency. Neither the European Union nor the granting authority can be held responsible for them. 
\end{acknowledgments}

% \section{Appendices}

%% The declaration on generative AI comes in effect
%% in Janary 2025. See also
%% https://ceur-ws.org/GenAI/Policy.html
\section*{Declaration on Generative AI}
 %  {\em Either:}\newline
 %  The author(s) have not employed any Generative AI tools.
 %  \newline
  
 % \noindent{\em Or (by using the activity taxonomy in ceur-ws.org/genai-tax.html):\newline}
 During the preparation of this work, the authors used ChatGPT in order for grammar and spelling checking, paraphrasing and rewording.
 % Further, the author(s) used X-AI-IMG for figures 3 and 4 in order to: Generate images. 
 After using these tool, the authors reviewed and edited the content as needed and take full responsibility for the publication’s content. 

%%
%% Define the bibliography file to be used
\bibliography{bibliography}

\end{document}